\documentclass[10pt,twocolumn,letterpaper]{article}

\usepackage{cvpr}              

\usepackage{graphicx}
\usepackage{amsmath}
\usepackage{amssymb}
\usepackage{booktabs}
\usepackage{multirow}
\usepackage{float}

\usepackage[pagebackref,breaklinks,colorlinks]{hyperref}

\usepackage[capitalize]{cleveref}
\crefname{section}{Sec.}{Secs.}
\Crefname{section}{Section}{Sections}
\Crefname{table}{Table}{Tables}
\crefname{table}{Tab.}{Tabs.}

\def\cvprPaperID{3} 
\def\confName{CVPR}
\def\confYear{2023}

\begin{document}

\title{Post-training Model Quantization Using GANs for Synthetic Data Generation}

\author{Athanasios Masouris\\
Delft University of Technology\\
Delft, The Netherlands\\
{\tt\small a.masouris@student.tudelft.nl}
\and
Mansi Sharma\\
Intel\\
Washington, USA\\
{\tt\small mansi.sharma@intel.com}
\and
Adrian Boguszewski \\
Intel\\
Swindon, UK\\
{\tt\small  adrian.boguszewski@intel.com}
\and
Alexander Kozlov \\
Intel\\
Dubai, UAE\\
{\tt\small alexander.kozlov@intel.com}
\and
Zhuo Wu \\
Intel\\
Shanghai, China\\
{\tt\small zhuo.wu@intel.com}
\and
Raymond Lo \\
Intel\\
Santa Clara, USA\\
{\tt\small raymond.lo@intel.com }
}

\maketitle

\begin{abstract}
Quantization is a widely adopted technique for deep neural networks to reduce the memory and computational resources required. However, when quantized, most models would need a suitable calibration process to keep their performance intact, which requires data from the target domain, such as a fraction of the dataset used in model training and model validation (\ie calibration dataset). 

In this study, we investigate the use of synthetic data as a substitute for the calibration with real data for the quantization method. We propose a data generation method based on Generative Adversarial Networks that are trained prior to the model quantization step. We compare the performance of models quantized using data generated by StyleGAN2-ADA and our pre-trained DiStyleGAN, with quantization using real data and an alternative data generation method based on fractal images. Overall, the results of our experiments demonstrate the potential of leveraging synthetic data for calibration during the quantization process. In our experiments, the percentage of accuracy degradation of the selected models was less than $0.6\%$, with our best performance achieved on \textit{MobileNetV2} ($0.05\%$). The code is available at: \url{https://github.com/ThanosM97/gsoc2022-openvino}

\end{abstract}

\section{Introduction}
\label{sec:intro}

Model optimization is essential in deep learning since it demonstrates efficacy in using computational resources, scalability, and even better performance. Techniques such as model pruning \cite{Han15pruning, he2017channel, srinivas2015data}, weight compression \cite{Han15compression}, and knowledge distillation \cite{ba2014deep, hinton2015distilling} have been proven to reduce the model complexity without sacrificing its accuracy.

Another approach that stands out is model quantization. Model quantization is a method used in deep learning to reduce a model's memory footprint and computational complexity and enable inference on hardware with limited resources (\eg smartphones, IoT devices). Additionally, quantization facilitates faster inference by reducing the computations required during forward propagation, and thus less power consumption.

Quantization is achieved by reducing the precision of model parameters and activations from their original floating-point precision to a lower-precision representation (\eg 8-bit fixed-point). There are two types of quantization, weight and activation. The goal of the weight quantization process is to find the nearest low-precision weights to the original floating-point ones. Activation quantization reduces the precision of the activations produced by the model. It can be achieved by either setting a precision based on the range of the activations during runtime (\ie dynamic fixed-point quantization) or by setting the same precision for all activations (\ie uniform fixed-point quantization). 

While a model can be quantized and trained from scratch (\ie \textit{quantization-aware training}), a more practical approach is \textit{post-training quantization}, which optimizes the performance of existing pre-trained models. However, post-training quantization can have a significant impact on model performance. The goal is to minimize the accuracy loss caused by quantization errors, while maximizing the efficiency of the end model. Specific methods, such as accuracy-aware quantization, can also be applied. In keeping the model's accuracy intact, these methods require a calibration dataset (\ie a small set of samples representative of the overall dataset the model was trained on). Nevertheless, due to privacy concerns, private use, or the scale of the dataset, a calibration subset may not always be available. In these cases, synthetic-generated data can be leveraged.

In this study, we investigate the use of synthetic data generated by Generative Adversarial Networks (GANs) \cite{goodfellow2020generative} as a substitute for the calibration with real data by the quantization method. GANs are known for their ability to generate new, realistic, and diverse data samples that are similar to the original data distribution. While the publication of a dataset might be hindered by privacy concerns or the sheer size of the data, a GAN model can become publicly available and thus be used to produce synthetic samples according to the users' needs or the requirements of a process (\ie quantization).

\section{Related Work}
\label{sec:related-work}
Prompted by the ever-increasing size of deep neural networks, techniques for reducing their computational costs have been thoroughly studied for efficient learning. Early attempts focused on reducing the number of network parameters by grouping the weights \cite{chen2015compressing}, replacing costly operations such as fully connected layers \cite{szegedy2015going}, or by pruning connections between layers \cite{lecun1989optimal, hassibi1993optimal, li2016pruning}. Network quantization has also been studied, with early work representing weights and activations using only a single bit, introducing the Binarized Neural Networks (BNNs) \cite{courbariaux2015binaryconnect, hubara2016binarized}. While this representation achieved a substantial reduction in computational costs, it also led to accuracy degradation on more complex models and datasets. In Gupta \etal \cite{gupta2015deep}, the authors demonstrated that using 16-bit fixed-point representation with stochastic rounding when training a CNN leads to negligible degradation in the classification accuracy. In Banner \etal \cite{banner2018scalable}, the precision was further reduced to an 8-bit representation while they quantized both weights and activations in all layers. Their proposed 8-bit training did not affect the models' accuracy when trained on a large-scale dataset. In Zhang \etal \cite{zhang2018lq}, the authors proposed a quantization method by training quantizers from data. In Han \etal \cite{Han15pruning}, they attempted to compress deep neural networks by combining pruning, trained quantization, and Huffman coding. Although their method could significantly compress the size of deep neural networks, the pruning process can be time-consuming and difficult to optimize.

Post-training quantization has also been the focus of research. In Lin \etal \cite{lin2016fixed}, the authors proposed an SQNR-based optimization approach to convert a pre-trained floating point deep convolutional model to its fixed-point equivalent. In Banner \etal \cite{banner2019post}, 4-bit quantization was proposed using the ACIQ method to optimize the clipping value, while in Choukroun \etal \cite{choukroun2019low}, they minimized the quantization MSE for both weights and activations. More recent studies suggested quantization by splitting outlier channels \cite{zhao2019improving}, using adaptive weight rounding \cite{nagel2020up}, or bias correction \cite{finkelstein2019fighting}. Additionally, given that post-training quantization requires a small calibration dataset, which may not be readily available, studies have addressed this issue by either employing data-free quantization techniques \cite{nagel2019data, cai2020zeroq} or using synthetic data \cite{lazarevich2021post}.

\section{Methodology}
\label{sec:methodology}

In this study, we examine the effect of the quantization process on the performance of the models in terms of classification accuracy. To do so, we utilized \textit{OpenVINO's Post-training Optimization Tool (POT)} \cite{POT}, which supports uniform integer quantization. Using this tool, the weights and activations of the classification models were converted from floating-point to integer precision (8-bit representation).

POT provides two quantization methods. First, the \textit{Default Quantization} process performs the quantization using a non-labeled calibration dataset to estimate the range of activation values. Second, the \textit{Accuracy-aware Quantization} provides control over the defined accuracy metric, allowing the tool to quantize specific layers of the model while maintaining the accuracy within a predefined range. In contrast with the default quantization, the accuracy-aware quantization requires a labeled calibration dataset.

Given that a calibration dataset may not always be available, in this study, we also investigate the use of synthetic data for the calibration process. We generated synthetic data using Generative Adversarial Networks (GANs) \cite{goodfellow2020generative}, StyleGAN2-ADA \cite{karras2020training} and our own DiStyleGAN (\cref{sec:distylegan}), pre-trained to approximate the distribution of the real data the classification models were trained on.

Subsequently, experiments were conducted using the Default Quantization and the Accuracy-aware quantization methods provided by POT. The quantized models were pre-trained on the classification task on the CIFAR-10 dataset \cite{krizhevsky2009learning}, and their performances upon quantization were compared on the CIFAR-10 test set. Furthermore, both quantization techniques were applied using multiple calibration datasets (real, synthetic, and fractal).

\subsection{Models}
Five models were selected to be quantized during the experiments. All of them were pre-trained on the CIFAR-10 dataset for classification in PyTorch and were obtained from PyTorch Hub\footnote{https://github.com/chenyaofo/pytorch-cifar-models}. The models, along with their corresponding versions, that were quantized were the following: \emph{ResNet20 (resnet20)} \cite{he2016deep}, \emph{VGG16 (vgg16\_bn)} \cite{simonyan2014very}, \emph{MobileNetV2 (mobilenetv2\_x1\_4)} \cite{sandler2018mobilenetv2}, \emph{ShuffleNetV2 (shufflenetv2\_x2\_0)} \cite{ma2018shufflenet}, and \emph{RepVGG (repvgg\_a2)} \cite{ding2021repvgg}.

\begin{figure*}[t]
  \centering
  \includegraphics[width=\textwidth]{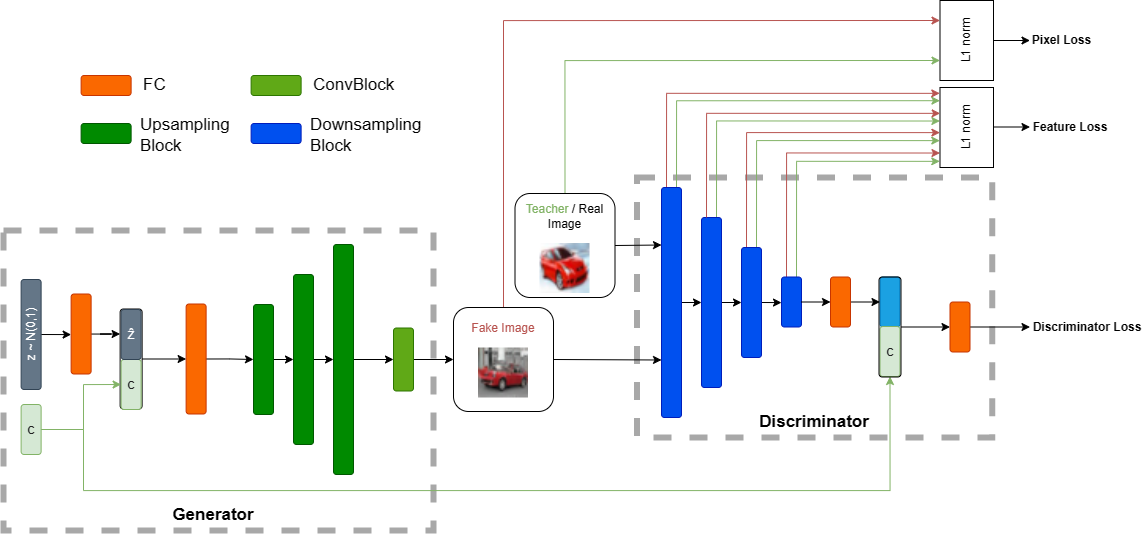}
  \caption{Overview of DiStyleGAN’s architecture}
  \label{fig:distylegan-arch}  
\end{figure*}

\subsection{GAN Training}
\label{sec:distylegan}
The current state-of-the-art model in class-conditional image generation on CIFAR-10 is StyleGAN2 \cite{Karras2019stylegan2}. While this model can generate synthetic images that look similar to the ones from the CIFAR-10 dataset, generating those images requires substantial computational resources due to its more than 20 million parameters. In an attempt to lower this requirement, we used the knowledge distillation framework described in Chang and Lu \cite{chang2020tinygan}. In particular, we used StyleGAN2 as a teacher network, to train a student network, DiStyleGAN (\cref{fig:distylegan-arch}), to produce synthetic images that approximate the CIFAR-10 distribution while requiring only about a tenth of the former's parameters. 

\subsubsection{Black-Box Distillation}
Similar to Chang \etal \cite{chang2020tinygan}, the teacher network is applied as a black box, requiring only limited access to its input-output pairs. In particular, $50,000$ synthetic images generated by StyleGAN2 were collecte, equally distributed among the $10$ classes of CIFAR-10, along with the input noise vectors and the corresponding class labels. This collection, or dataset, is then used to train DiStyleGAN in a supervised way. Following this approach, no knowledge of the internal or intermediate features of the teacher model is required and can be discarded afterward.

\subsubsection{Objectives}
For the training of DiStyleGAN, we leveraged the same objectives as in \cite{chang2020tinygan}.

\begin{equation}
\mathcal{L}_G = \mathcal{L}_{KD_{feat}} + \lambda_1 \mathcal{L}_{KD_{pix}} + \lambda_2 \mathcal{L}_{KD_S} + \lambda_3 \mathcal{L}_{GAN_S}    
\end{equation}

\noindent where $\mathcal{L}_{KD_{feat}}$ is the feature-level distillation loss calculated from feature maps extracted using the Discriminator's network for the images generated by both the student and the teacher networks, $\mathcal{L}_{KD_{pix}}$ is the pixel-level distillation loss calculated as the pixel $L_1$ distance between the student and teacher images, $\mathcal{L}_{KD_S}$ is the adversarial distillation loss used to train the student generator to approximate the distribution of the teacher's, and $\mathcal{L}_{GAN_S}$ is the adversarial GAN loss used to train the student generator to approximate the distribution of the real data.

For the discriminator, the following objective was used:

\begin{equation}
\mathcal{L}_D = \mathcal{L}_{KD_D} + \lambda_4 \mathcal{L}_{GAN_D}    
\end{equation}

\noindent where $\mathcal{L}_{KD_D}$ is the adversarial distillation loss used to encourage the discriminator to distinguish between generated images by the student and teacher networks and $\mathcal{L}_{GAN_D}$ is the adversarial GAN loss used to encourage discrimination between student and real images.

\begin{table*}[t]
\centering
\subfloat[Default Quantization algorithm. Even if the best results are achieved with the original test set, the accuracy degradation is minimal (less than 1.5\%) for most cases using synthetic data.\label{tab:def-quant-acc-drop}]{
\resizebox{0.61\textwidth}{!}{
\begin{tabular}{cccccc}
\toprule
\multirow{2}{*}{Calibration Dataset} & \multicolumn{5}{c}{Model}                                             \\ 
                                     & ResNet20 & VGG16  & MobileNetV2 & ShuffleNetV2     & RepVGG          \\
\midrule
CIFAR-10                             & \textbf{0.28\%}   & \textbf{0.05\%} & 0.07\%      & \textbf{1.26\%}           & 41.94\% \\
StyleGAN2-ADA                        & 0.43\%   & 0.16\% & \textbf{0.05\%}      & 7.48\%  & \textbf{41.37\%} \\
DiStyleGAN                           & 0.46\%   & 0.07\% & 0.16\%      & 1.49\%           & 42.85\% \\
Fractal                              & 1.10\%   & 0.69\% & 0.54\%      & 87.21\% & 41.73\% \\
\bottomrule
\end{tabular}
}
}
\hfill
\subfloat[Accuracy-aware Quantization algorithm. The results from synthetic data (except fractal images) are comparable to the real  dataset.\label{tab:acc-quant-acc-drop}]{
\resizebox{0.345\textwidth}{!}{
\begin{tabular}{ccc}
\toprule
\multirow{2}{*}{Calibration Dataset} & \multicolumn{2}{c}{Model} \\
                                     & ShuffleNetV2   & RepVGG   \\
\midrule
CIFAR-10                             & 1.26\%         & \textbf{0.45\%}   \\
StyleGAN2-ADA                        & 0.59\%         & 0.48\%   \\
DiStyleGAN                           & \textbf{0.11\%}         & 0.55\%   \\
Fractal                              & 87.21\%        & 42.16\% \\
\bottomrule
\end{tabular}
}
}
\caption{Accuracy drop for each model and calibration dataset, as measured in the classification task on the CIFAR-10 test set.}
\end{table*}

\subsubsection{Network Architectures}
\textbf{Generator} 
Initially, the Gaussian random noise vector is projected to 128 dimensions using a Fully Connected layer. Subsequently, the condition embedding and a projected noise vector are concatenated and passed through another Fully Connected layer, followed by three consecutive Upsampling blocks. Each upsampling block consists of an upsample layer (scale\_factor=2, mode=’nearest’), a 3x3 convolution with padding, a Batch Normalization layer, and a Gated Linear Unit (GLU). Finally, there is a convolutional block consisting of a 3x3 convolution with padding and a hyperbolic tangent activation function (tanh), which produces the fake image.

\textbf{Discriminator}
DiStyleGAN’s discriminator consists of 4 consecutive Downsampling blocks (4x4 strided-convolution, Spectral Normalization, and a LeakyReLU), with each of them reducing the spatial size of the input image by a factor of 2. These four blocks also produce the feature maps that calculate the Feature Loss. Subsequently, the logit is flattened, projected to 128 dimensions, and concatenated with the class condition embedding before being passed through a final fully connected layer to produce the class-conditional discriminator loss.

\subsection{Calibration Datasets}
\label{sec:methodology-calibration-datasets}
The data used for the calibration of the quantization process came from four different distributions. First, we conducted quantization using a subset of the real CIFAR-10 data as a frame of reference. Second, we opted for synthetic images produced by the StyleGAN2-ADA model, the state-of-the-art model in class-conditional image generation on CIFAR-10. Third, we produced synthetic images using DiStyleGAN. Finally, fractal image data generated by Datumaro \cite{Datumaro} were also used. While the former two synthetic datasets approximate the CIFAR-10 distribution, and thus could be considered representative, the fractal images do not constitute a representative dataset for the deep learning models pre-trained on CIFAR-10. However, we include them in our experiments since Lazarevich \etal \cite{lazarevich2021post} demonstrated that it is possible to perform post-training quantization even with a non-representative dataset. 

\begin{figure*}[t]
  \centering
  \includegraphics[width=\textwidth]{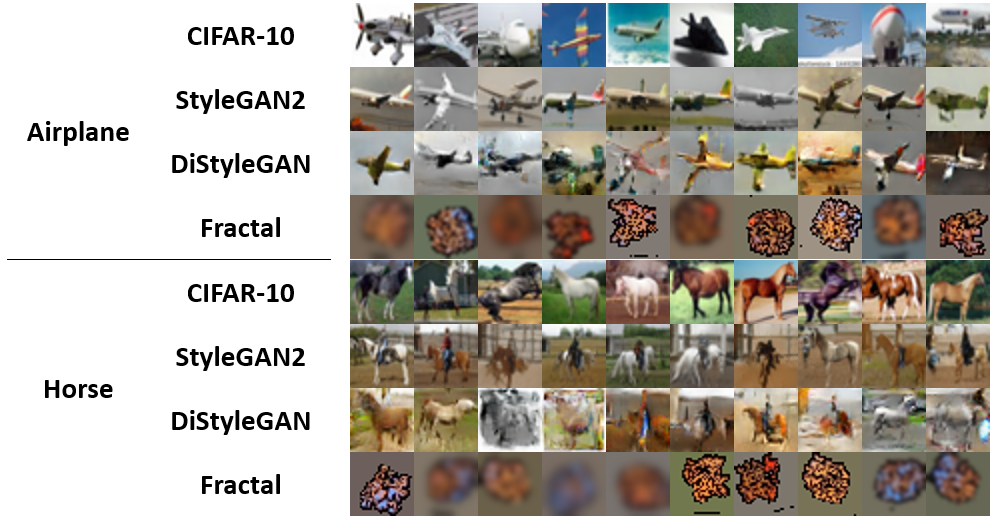}
  \caption{Samples from each of the four calibration datasets for the CIFAR-10 classes "Airplane" and "Horse".}
  \label{fig:cifar-samples}  
\end{figure*}

Samples from each of the four data distributions for the classes \textit{"Airplane``} and \textit{"Horse``} are illustrated in \Cref{fig:cifar-samples}, while samples for the rest of the classes can be found in \cref{apdx:A}. Additionally, \Cref{tab:inception-score} presents the \textit{Inception Score (IS)} \cite{salimans2016improved} calculated for each data distribution. The inception score is a metric that evaluates the quality of synthetic generated images. The calculated score is based on the ability of a pre-trained InceptionV3 \cite{szegedy2015rethinking} model to classify the images of a synthetic dataset produced by a generative model. In our case, we calculated the inception scores using datasets of $50,000$ synthetic samples from each distribution. 

\begin{table}[ht]
\label{tab:inception-score}
\centering
\begin{tabular}{lc}
\toprule
Model         & Inception Score \\
\midrule
StyleGAN2-ADA & 10.34           \\
DiStyleGAN    & 6.78            \\
Fractal       & 3.31\\
\bottomrule
\end{tabular}
\caption{Inception score calculated for each synthetic dataset used in the calibration process (higher is better).}
\end{table}

For our quantization experiments, $5000$ images were used from each of the aforementioned datasets ($500$ images per class of the CIFAR-10 dataset). Then, the official CIFAR-10 test set was utilized to evaluate the results of the quantized models for all combinations of calibration datasets, PyTorch models, and quantization techniques.


\section{Experiments}
\subsection{Results}

\Cref{tab:def-quant-acc-drop,tab:acc-quant-acc-drop} showcase the accuracy degradation percentage of the quantized models with respect to the performance of the original PyTorch models on the classification task on the CIFAR-10 test set. These results were obtained using the two different quantization algorithms, default and accuracy-aware, and the four calibration datasets described in \Cref{sec:methodology-calibration-datasets}. In addition,  \Cref{tab:inception-score} presents the inception scores of the synthetic calibration datasets, a quantitative metric that indicates the difference in the quality of the generated images between the four aforementioned datasets. 

\subsection{Discussion}
\label{sec:discussion}

Based on the results obtained by the experiments, it is noticeable that in most cases, the Default Quantization algorithm can successfully quantize the models while showing only a negligible degradation in accuracy. This especially holds true for the ResNet20, VGG, and MobileNetV2 networks. Even with a non-representative calibration dataset (i.e. Fractal), the accuracy decrease is limited to a maximum of 1.1\%, as showcased in \Cref{tab:def-quant-acc-drop}, which is in par with the findings of Lazarevich \etal \cite{lazarevich2021post}.

When the default quantization algorithm fails, the accuracy-aware algorithm can be employed. The RepVGG model suffers the most by the default algorithm, with more than a 40\% accuracy decrease across all calibration datasets. This could be potentially attributed to the high variance of RepVGG's activation weights, constituting the weight distribution quite different across different channels. However, further experimentation is required to validate this claim. On the other hand, the results of the accuracy-aware quantization shown in \Cref{tab:acc-quant-acc-drop} prove that the same model can be quantized using synthetic data with minimal accuracy degradation ($\approx$ 0.5\%). The same, but less significant, effect can be observed for the ShuffleNetV2 model. It should also be noted that the accuracy-aware quantization requires a representative calibration dataset; thus the corresponding results for the fractal images were expected.

Finally, it is essential to notice that the two synthetic datasets, StyleGAN2-ADA and DiStyleGAN, lead to comparable results with the official CIFAR-10 dataset in the case of quantization. Surprisingly enough, although the DiStyleGAN model was trained through knowledge distillation, with the StyleGAN2-ADA as the teacher network, there are cases where the quantization process using the dataset generated by the former leads to better results compared to when the latter is used. This finding is also in contrast with the quantitative results for the synthetic generated images (\cref{tab:inception-score}) which clearly showcase the superiority of the StyleGAN2-ADA model compared to our DiStyleGAN, in terms of the quality of their synthetic generated images. However, the inception score metric does not take into consideration how synthetic images compare to real images. Other metrics (\eg \textit{Fréchet inception distance (FID)} \cite{heusel2017gans}) that evaluate the synthetic images while taking into account the distribution of the real data, should also be included in further experimentation. 

\section{Conclusion}

Based on our experiments, post-training quantization leads to minimal accuracy degradation for three of the selected models (ResNet20, VGG16, MobileNetV2), regardless of the used calibration dataset. For the remaining models (ShuffleNetV2, RepVGG), while an accuracy-aware quantization is required, the results indicate similar performance when the real data are used for calibration compared to the synthetic. Insignificant differences were also observed between the quantized models, when synthetic data produced by the complex StyleGAN model were used, compared to when they were produced by DiStyleGAN. This finding suggests that in the case of quantization when the synthetic data of the calibration dataset approximate the distribution of the real data, even a simple generator might be enough. However, further experiments with synthetic calibration datasets of varied quality are required to corroborate this finding. 


{\small
\bibliographystyle{ieee_fullname}
\bibliography{PaperForReview}
}

\clearpage
\appendix
\counterwithin{figure}{section}

\onecolumn
\section{Calibration Datasets Samples}
\label{apdx:A}

\begin{figure*}[!ht]
  \centering
  \includegraphics{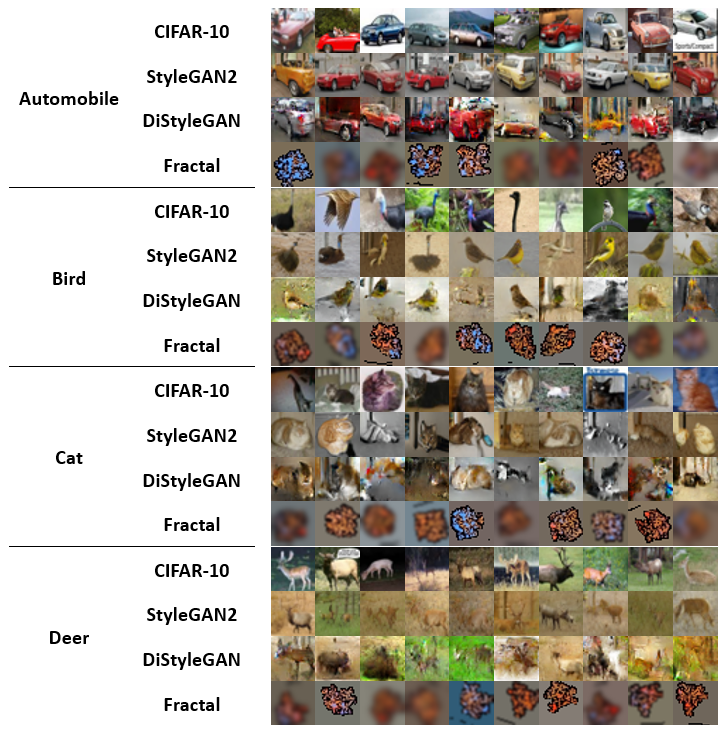}
  \caption{Samples from each of the four calibration datasets for the CIFAR-10 classes: "Automobile", "Bird", "Cat", and "Deer``.}
\end{figure*}

\begin{figure*}[!ht]
  \centering
  \includegraphics{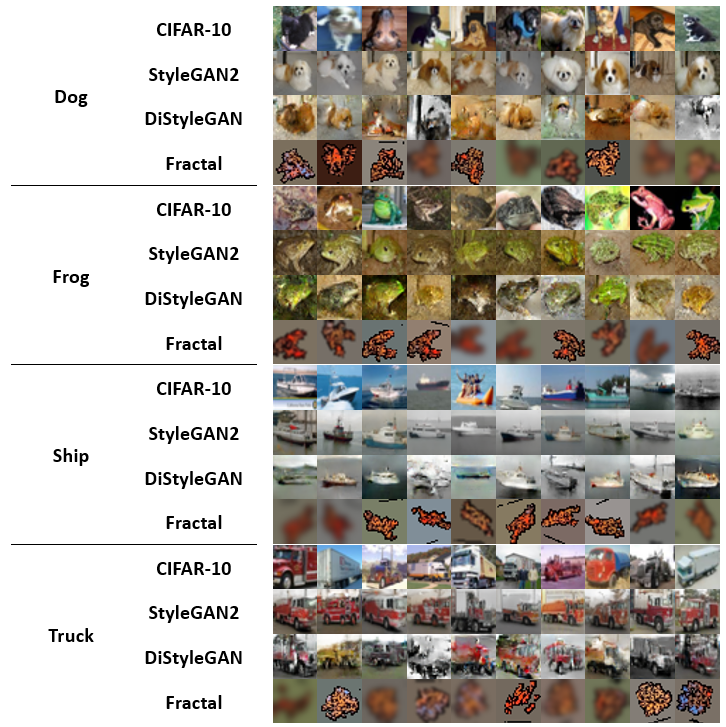}
  \caption{Samples from each of the four calibration datasets for the CIFAR-10 classes: "Dog", "Frog", "Ship", and "Truck".}
  \label{fig:cifar-samples-3}  
\end{figure*}

\end{document}